\documentclass[11pt]{article}
\usepackage{semdial2018}
\usepackage{times}
\usepackage{url}
\usepackage{latexsym}

\usepackage{amsmath}
\usepackage{graphicx}
\usepackage[colorinlistoftodos]{todonotes}

\usepackage{lingmacros}
\usepackage{ds}
\usepackage{ttr}

\title{Exploring Semantic Incrementality \\ with Dynamic Syntax and Vector Space Semantics}

\author{Mehrnoosh Sadrzadeh,$^{1}$ Matthew Purver,$^{1}$ Julian Hough,$^{1}$ Ruth Kempson$^{2}$\\
$^{1}$School of Electronic Engineering and Computer Science,\\
Queen Mary University of London\\
{\tt \{m.sadrzadeh,m.purver,j.hough\}@qmul.ac.uk}\\
$^{2}$Department of Philosophy, King's College London\\
{\tt ruth.kempson@kcl.ac.uk}
}
\date{}

\begin{document}
\maketitle
\begin{abstract}
One of the fundamental requirements for models of semantic processing in dialogue is \emph{incrementality}: a model must reflect how people interpret and generate language at least on a word-by-word basis, and handle phenomena such as fragments, incomplete and jointly-produced utterances. We show that the incremental word-by-word parsing process of Dynamic Syntax (DS) can be assigned a compositional distributional semantics, with the composition operator of DS corresponding to the general operation of tensor contraction from multilinear algebra. We provide abstract semantic decorations for the nodes of DS trees,  in terms of vectors, tensors, and sums thereof; using the latter to model the underspecified elements crucial to assigning partial representations during incremental processing. As a working example, we  give an instantiation of this theory using plausibility tensors of compositional distributional semantics, and show how our framework can incrementally assign a semantic plausibility measure as it parses phrases and sentences. 
\end{abstract}

%
%
\blfootnote{
    %
    %
    %
    %
    \hspace{-0.65cm}  
    This work is licenced under a Creative Commons 
    Attribution 4.0 International Licence.
    Licence details:
    \url{http://creativecommons.org/licenses/by/4.0/}
    %
    %
}
\section{Introduction}

An incremental, word-by-word view on language processing is motivated by much empirical evidence from human-human dialogue. This evidence includes split, interrupted, and corrective utterances, see e.g.\ \cite{Howes.etal11}:

\vspace*{-0.1cm}
\enumsentence{
  A:  Ray destroyed  $\ldots$ \\
  B:  $\ldots$ the fuchsia. He never liked it.  The roses he spared $\ldots$ \\
  A: $\ldots$ this time.
}\label{ex:fuchsia}
\vspace*{-0.1cm}


\noindent
In (\ref{ex:fuchsia}), the utterances are either inherently incomplete or potentially complete, with more than one agent contributing to the unfolding of a sequence, with in principle arbitrary speaker switch points and indefinite extendibility.  In such cases, speakers and hearers must be processing the structural and semantic information encoded in each utterance incrementally. A second motivation comes from computational dialogue systems, where the ability to process incrementally helps speed up systems and provide more natural interaction \cite{Aist.etal07}.
 A third motivation comes from psycholinguistic results, even in individual language processing, which show that hearers can incrementally disambiguate word senses and resolve references, before sentences are complete and even using partial words and disfluent material to do so \cite{Brennan.Schober01}. In (\ref{ex:dribble}a,b), the ambiguous word \textit{dribbled}  can be resolved to a particular sense early on, given the (\textit{footballer} or \textit{baby}) subject, without waiting for the rest of the sentence. A fourth comes from cognitive neuroscience and models such as Predictive Processing \cite{FristonFrith15duet,Clark15surfing} which focus on agents' incremental ability to generate expectations and judge the degree to which they are met by observed input.

\eenumsentence{
\item The footballer dribbled $\ldots$ ~ $\ldots$ the ball across the pitch.
\vspace*{-0.25cm}
\item The baby dribbled $\ldots$ ~ $\ldots$ the milk all over the floor.
}\label{ex:dribble}


We use the framework of Dynamic Syntax (DS) for incremental grammatical and semantic analysis \cite{Kempson.etal01,Cann.etal05,Kempson.etal16TheoreticalLing}. DS has sufficient expressivity to capture the dialogue phenomena in (\ref{ex:fuchsia}) and has been used to provide incremental interpretation and generation for dialogue systems \cite{PurverEshghiHough2011,eshghi-shalyminov-lemon:2017:EMNLP2017}. Yet incremental disambiguation is currently beyond its expressive power; and while its framework is broadly predictive, it does not yet provide an explanation for how specific expectations can be generated or their similarity to observations measured- though see \cite{HoughPurver17Lattices} for DS's interface to a probabilistic semantics.

DS does not fix a special form of syntax and instead defines grammaticality directly in terms of incremental semantic tree growth. Symbolic methods are employed for labelling the contents of these trees,  via terms either from an epsilon calculus \cite{Kempson.etal01} or a suitable type theory with records \cite{Purver.etal10}.  These symbolic approaches are not easily able to reflect the non-deterministic content of  natural language forms, nor the way any  initially unfixable  interpretation, polysemy being rampant,  can be narrowed down during the utterance interpretation process.  For the same reason, the assigned term specifications do not  provide a basis for the graded judgements that humans are able to make during processing to assess similarity to (or divergence from) expectations \cite{Clark15surfing},  to incrementally narrow down a word's interpretation, or disambiguate its sense in the emerging context. 

Non-determinisms of meaning and gradient similarity judgements are the stronghold of the so-called distributional or vector space semantics \cite{Salton,Schutze,Lin,Curran}. 
By modelling word meanings as vectors within a continuous space, such approaches  directly express graded similarity of meaning (e.g.\ as distance or angle between vectors) and changes in interpretation (via movements of vectors within a space).
%
%
 %
Vector space semantics has been extended from words to phrases and sentences using  different grammatical formalisms, e.g. Lambek pregroups, Lambek Calculus, and Combinatory Categorial Grammar (CCG)  \cite{Maillard,krishnamurthy,Coeckeetal,CoeckeGrefenSadr}.  It has, however, not been extended to incremental and dialogue formalisms such as DS.


In this paper, we address these lacunae, by defining an incremental  vector space semantic model  for DS that can express non-determinism and similarity in word meaning, and yet keep incremental compositionality over conversational exchanges. 
As a working example, we  instantiate this model using the plausibility instance of \cite{SClark15} developed for a type-driven compositional distributional semantics, and show how it can  incrementally assign a semantic plausibility measure as it performs  word-by-word parses of  phrases and sentences. We  discuss how this ability enables us to  incrementally disambiguate words using their immediate contexts and   to model the plausibility of continuations and thus a hearer's expectations.


\section{Dynamic Syntax and its Semantics}
In its original form, Dynamic Syntax (DS) provides a strictly incremental formalism relating word sequences to semantic representations. Conventionally, these are seen as trees decorated with semantic formulae that are terms in a typed lambda calculus \cite{Kempson.etal01}, chapter 9:

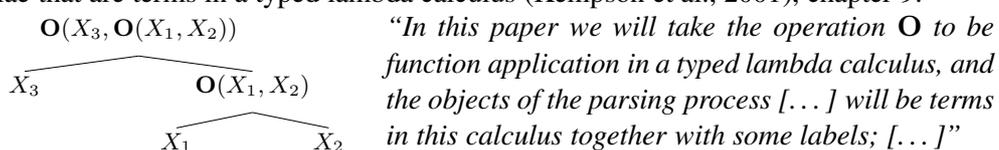
\begin{tabular}{cp{8cm}}
\dstree{0.75}{2}{
\br{$\mathbf{O}(X_3,\mathbf{O}(X_1,X_2))$}{
\lf{$X_3$} \br{$\mathbf{O}(X_1,X_2)$}{\lf{$X_1$} \lf{$X_2$}}
}
}
&
\it ``In this paper we will take the operation $\mathbf{O}$ to be \textit{function application} in a typed lambda calculus, and the \textit{objects} of the parsing process [\ldots] 
will be terms in this calculus together with some labels; [\ldots]" 
\end{tabular}

\begin{figure}[!ht]
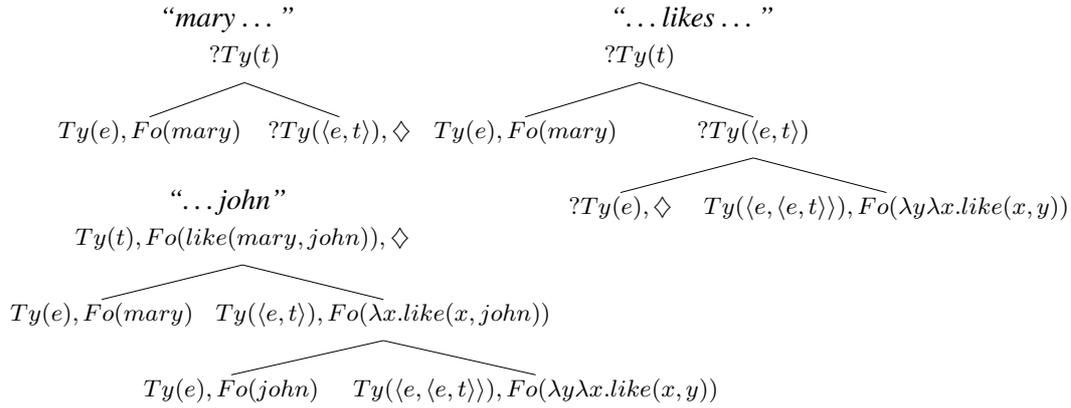

\begin{tabular}{cc}
& \vspace*{-0.25cm}\\
\hspace*{-2cm}
\textit{``mary \ldots''}
& 
\hspace*{-2cm}
\textit{``\ldots likes \ldots''}

\\

\hspace*{-2cm}
\dstree{1}{2.5}{
\br{$?Ty(t)$}{
 \lf{$Ty(e),Fo(mary)$} 
 \lf{$?Ty(\langle e,t\rangle),\diamondsuit$}
 }
}
&
\hspace*{-2cm}
\dstree{1}{1.25}{
\br{$?Ty(t)$}{
 \lf{$Ty(e),Fo(mary)$} 
 \psset{levelsep=1cm, treesep=3.5cm}
 \br{$?Ty(\langle e,t\rangle)$}{
  \lf{$?Ty(e),\diamondsuit$}
  \lf{$Ty(\langle e,\langle e,t\rangle\rangle),Fo(\lambda y\lambda x.like(x,y))$}
  }
 }
}
\vspace*{-0.5cm}
\\

\hspace*{-2cm}
\textit{``\ldots john''}
& \\
\dstree{1}{1.5}{
\br{$Ty(t),Fo(like(mary,john)),\diamondsuit$}{
 \lf{$Ty(e),Fo(mary)$} 
 \psset{levelsep=1cm, treesep=4cm}
 \br{$Ty(\langle e,t\rangle),Fo(\lambda x.like(x,john))$}{
  \lf{$Ty(e),Fo(john)$}
  \lf{$Ty(\langle e,\langle e,t\rangle\rangle),Fo(\lambda y\lambda x.like(x,y))$}
  }
 }
}
& \vspace*{0.25cm}\\
\end{tabular}
\caption{DS parsing as semantic tree development, for the simple sentence \textit{``mary likes john''}.}\label{fig:mlj}
\end{figure}

\noindent
This allows us to give analyses of the semantic output of the word-by-word parsing process in terms of partial semantic trees, in which nodes are labelled with type $Ty$ and semantic formula $Fo$, or with requirements for future development (e.g.\ $?Ty$. $?Fo$), and with a pointer $\diamondsuit$ indicating the node currently under development. This is shown in Figure~\ref{fig:mlj} for the simple sentence \textit{Mary likes John}.
Phenomena such as conjunction, apposition and relative clauses are analysed via  {\sc Link}ed trees (corresponding to semantic conjunction). For reasons of space we do not present an original DS tree here;  an example of a non-restrictive relative clause linked tree labelled with  vectors is presented in  Figure~\ref{fig:linkvector}.

\noindent
However, the DS formalism is in fact considerably more general. To continue the quotation above:

\begin{quote}
\it
``[\ldots] it is important to keep in mind that the choice of the actual representation language is not central to the parsing model developed here. [\ldots]
For instance, we may take $X_1, X_2, X_3$ to be feature structures and the operation $\mathbf{O}$ to be unification, or  $X_1, X_2, X_3$ to be lambda terms and $\mathbf{O}$ Application, or $X_1, X_2, X_3$ to be labelled categorial expressions and $\mathbf{O}$ Application: Modus Ponens, or $X_1, X_2, X_3$ to be DRSs and $\mathbf{O}$ Merging." 
\end{quote}

\noindent
Indeed, in some variants this generality is exploited; for example, \newcite{Purver.etal10} outline a version in which the formulae are \emph{record types} in Type Theory with Records (TTR) \cite{Cooper05}; and \newcite{Hough.Purver12} show how this can confer an extra advantage -- the incremental decoration of the \emph{root} node, even for partial trees, with a maximally specific formula via type inference, using the TTR merge operation $\ttrmerge$ as the composition function. In the latter account, underspecified record types decorate requirement nodes, containing a type judgement with the relevant type (e.g.\ $\smttrrectype{x & e}$ at type $?Ty(e)$ nodes). \newcite{HoughPurver17Lattices} show that this underspecification can be given a precise semantics through record type lattices: the dual operation of merge, the minimum common super type (or join) $\ttrminsuper$ is required to define a (probabilistic) distributive record type lattice bound by $\ttrmerge$ and $\ttrminsuper$. The interpretation process, including reference resolution, then takes the incrementally built top-level formula and checks it against a type system (corresponding to a world model) defined by a record type lattice. Implicitly, the record type on each node in a DS-TTR tree can be seen to correspond to a potential set of type judgements as sub-lattices of this lattice, with the appropriate underspecified record type  (e.g. $\smttrrectype{x & e}$) as their top element, with a probability value for each element in the probabilistic TTR version. In this paper, we show how equivalent underspecification, and narrowing down of meaning over time --- but with the additional advantages inherent in vector space models, e.g.\ similarity judgements --- can be defined for vector space representations with analogous operations to $\ttrmerge$ and $\ttrminsuper$.

\section{Compositional Vector Space Semantics for DS}


Vector space semantics are commonly instantiated via lexical co-occurrence, based on the \emph{distributional hypothesis} that meanings of words are represented by the distributions of the words around them- this is often described by Firth's claim that ``you shall know a word by the company it keeps'' \cite{Firth}. This can be implemented by creating a co-occurrence matrix \cite{Rubenstein}, whose columns are labelled by context words and whose rows by target words; the entry of the matrix at the intersection of a  context word $c$ and a target word $t$ is a function (such as TF-IDF or PPMI) of the number of times $t$ occurred in the context of $c$ (as defined via e.g.\ a lexical neighbourhood window, a dependency relation, etc.). The meaning of each target word is  represented by its corresponding row of the matrix. These rows are embedded in a vector space, where the  distances between the vectors represent  degrees of semantic similarity between  words \cite{Schutze,Lin,Curran}. 

Distributional semantics has  been extended from word level to sentence level, where a compositional operation acts on the vectors of the words to produce a vector for the sentence. Existing models vary from using simple additive and multiplicative compositional operations \cite{MitchellLapata} to compositional operators based on fully fledged categorial grammar derivations, e.g.\ pregroup grammars \cite{Coeckeetal,SClark15} or CCG \cite{krishnamurthy,Baroni,Maillard}. However, the work done so far has not been directly compatible with incremental processing: this paper is the first attempt to develop such an incremental semantics, using a framework not based on a categorial grammar, i.e. one in which a full categorial analysis of the phrase/sentence is not the obligatory starting point.

Compositional vector space semantic models have a  complementary property to DS. Whereas DS is agnostic to its choice of semantics,  compositional vector space models are agnostic to the choice of  the syntactic system. \newcite{Coeckeetal} show how they provide semantics for sentences based on the grammatical structures given by Lambek's pregroup grammars \cite{LambekPreg}; \newcite{CoeckeGrefenSadr} show how this semantics also works starting from the parse trees of Lambek's Syntactic Calculus \cite{LambekSyn}; \newcite{Wijnholds} shows how the same semantics can be extended to the Lambek-Grishin Calculus; and  \cite{krishnamurthy,Baroni,Maillard} show how  it  works for CCG trees. These semantic models homomorphically map the concatenation and slashes of categorial grammars to tensors and their evaluation/application/composition operations  to tensor contraction. 

In  DS terms, structures $X_1, X_2, X_3$ are mapped to general higher order tensors, e.g. as follows:   
\[
\begin{array}{cclcl}
X_1 &\mapsto &T_{i_1i_2 \cdots i_n} &\in& V_1 \otimes V_2 \otimes \cdots V_n\\ 
X_2 &\mapsto &T_{i_{n} i_{n+1} \cdots i_{n+k}}&\in &V_n \otimes V_{n+1} \otimes \cdots V_{n+k}\\  X_3 &\mapsto &T_{i_{n+k} i_{n+k+1} \cdots i_{n+k+m}} &\in &V_{n+k} \otimes V_{n+k+1} \otimes \cdots V_{n+k+m}
\end{array}
\]
Each  $T_{i_1i_2 \cdots i_n}$ abbreviates   the linear expansion of a tensor, which is normally written as follows:
\[
T_{i_1i_2 \cdots i_n} \equiv \sum_{i_1i_2 \cdots i_n} C_{i_1i_2 \cdots i_n} {e}_1 \otimes {e}_2 \otimes \cdots \otimes {e}_n
\]
for ${e}_i$  a basis of $V_i$ and $C_{i_1i_2 \cdots i_n}$ its corresponding scalar value. 
The $\mathbf{O}$ operations are mapped to contractions between these tensors, formed as follows:
\[
\begin{array}{ccl}
\mathbf{O}(X_1, X_2) &\mapsto &
T_{i_1 i_2 \cdots  i_n } T_{i_{n} i_{n+1} \cdots i_{n+k}}\\
&\in & V_1 \otimes V_2 \otimes \cdots \otimes V_{n-1} \otimes V_{n+1} \otimes \cdots \otimes V_{n+k}\\
\mathbf{O}(X_3, \mathbf{O}(X_1, X_2)) &\mapsto& T_{i_1 i_2 \cdots  i_n } T_{i_{n} i_{n+1} \cdots i_{n+k}}T_{i_{n+k} i_{n+k+1} \cdots i_{n+k+m}} \\
& \in & V_1 \otimes V_2 \otimes \cdots \otimes V_{n-1}\otimes V_{n+1} \otimes \cdots \otimes V_{n+k-1} \otimes V_{n+k+1} \otimes \cdots \otimes V_{n+k+m} 
\end{array}
\]
In their most general form presented above, these formulae are large  and the index notation becomes difficult to read. In special cases, however, it is often enough to work with spaces of rank around  3. For instance, the application of a transitive verb to its object is mapped to the following contraction: 
\[
T_{i_1 i_2  i_3 } T_{i_3} = (\sum_{i_1i_2i_3} C_{i_1i_2i_3} {e}_1 \otimes {e}_2 \otimes {e}_3)(\sum_{i_3} C_{i_3} {e}_3) = \sum_{i_1i_2} C_{i_1i_2i_3} C_{i_3}{e}_1 \otimes {e}_2
\]
This is the contraction between a cube $T_{i_1 i_2  i_3 }$ in $X_1 \otimes X_2 \otimes X_3$ and a vector $T_{i_3}$ in $X_3$,  resulting in a matrix in   $T_{i_1i_2}$ in $ X_{1} \otimes X_2$.

\begin{figure}[!t]
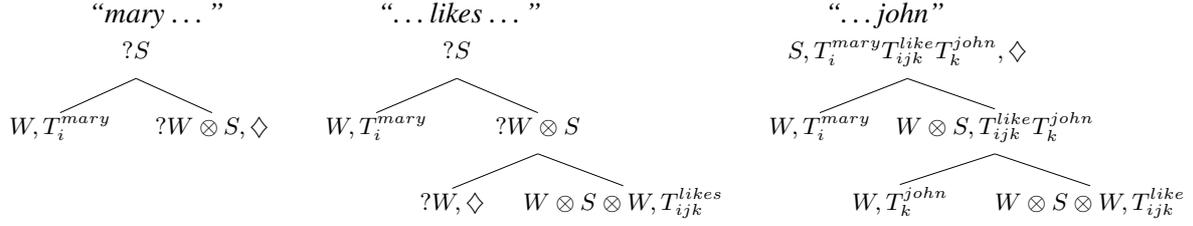

\begin{center}
\begin{tabular}{ccc}
& & \\
\hspace*{-0.5cm}
\textit{``mary \ldots''}
& 
\textit{``\ldots likes \ldots''}
&
\hspace*{-0.5cm}
\textit{``\ldots john''}

\\

\hspace*{-1.5cm}
\dstree{1}{2}{
\br{$?S$}{
 \lf{$W, T_i^{mary}$} 
 \lf{$?W \otimes S,\diamondsuit$}
 }
}
&
\hspace*{0.5cm}
\dstree{1}{1}{
\br{$?S$}{
 \lf{$W, T_i^{mary}$} 
 \psset{levelsep=1cm, treesep=2.25cm}
 \br{$?W \otimes S$}{
  \lf{$?W,\diamondsuit$}
  \lf{$W \otimes S \otimes W, T_{ijk}^{likes}$}
  }
 }
}


& 
\hspace*{0.5cm}
\dstree{1}{1}{
\br{$S , T_i^{mary}T_{ijk}^{like}T_k^{john},\diamondsuit$}{
 \lf{$W , T_i^{mary}$} 
 \psset{levelsep=1cm, treesep=2.5cm}
 \br{$W \otimes S , {T_{ijk}^{like}}T_k^{john}$}{
  \lf{$W, T_k^{john}$}
  \lf{$W \otimes S \otimes W , T_{ijk}^{like}$}
  }
 }
}
\\
& & \\

\end{tabular}
\end{center}
\caption{A DS with Vector Space Semantics parse of `Mary likes John'.}\label{fig:marylikesjohn}
\end{figure}

\begin{figure}[!b]
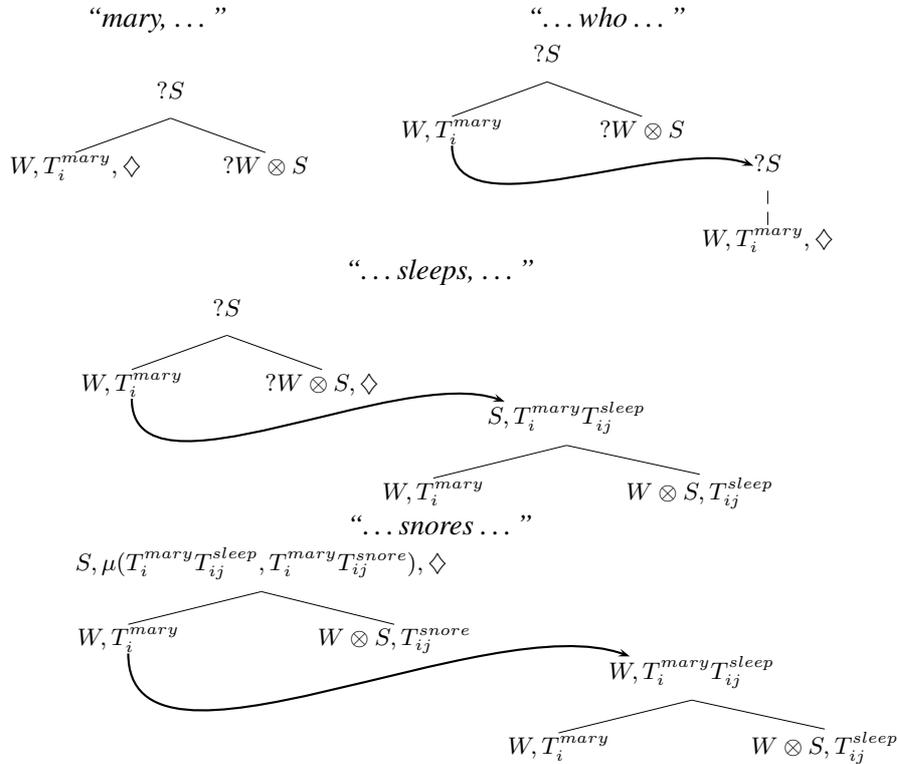

\begin{center}
\begin{tabular}{cc}
& \vspace*{-0.25cm}\\
\textit{``mary, \ldots''}
& 
\textit{``\ldots who \ldots''}

\\

\begin{tabular}{c}
\dstree{1}{2.5}{
\br{$?S$}{
 \lf{$W , T_i^{mary},\diamondsuit$} 
 \lf{$?W \otimes S$}
 }
}
\\
\\
\end{tabular}

&

\begin{tabular}{cc}
\dstree{1}{2.5}{
\br{$?S$}{
 \lf[M]{$W , T_i^{mary}$}
 \lf{$?W \otimes S$}
 }
} 
& \\
& 
\dstree{1}{1.5}{
\br[W]{$?S$}{
 \psset{linestyle=dashed}
 \lf{$W , T_i^{mary},\diamondsuit$}
 }
}
\nccurve[ncurv=.6,angleA=270,angleB=160]{->}{M}{W}
\\
\end{tabular}

\\

\multicolumn{2}{c}{\textit{``\ldots sleeps, \ldots''}}\\ 

\multicolumn{2}{c}{
\hspace*{-1cm}
\begin{tabular}{cc}
\dstree{1}{2.5}{
\br{$?S$}{
 \lf[M]{$W , T_i^{mary}$}
 \lf{$?W \otimes S,\diamondsuit$}
 }
} 
& \\
& 
\dstree{1}{3.5}{
\br[W]{$S , T_i^{mary}T_{ij}^{sleep}$}{
 \lf{$W,T_i^{mary}$}
 \lf{$W \otimes S , T_{ij}^{sleep}$}
 }
}
\nccurve[ncurv=.6,angleA=270,angleB=160]{->}{M}{W}
\\
\end{tabular}
}\\
\multicolumn{2}{c}{\textit{``\ldots snores \ldots''}}\\

\multicolumn{2}{c}{

\begin{tabular}{cc}
\dstree{1}{3.5}{
\br{$S, \mu(T_i^{mary}T_{ij}^{sleep},  T_i^{mary}T_{ij}^{snore}),\diamondsuit$}{
 \lf[M]{$W , T_i^{mary}$}
 \lf{$W \otimes S , T_{ij}^{snore}$}
 }
} 
& \\
& 
\dstree{1}{3.5}{
\br[W]{$W,T_i^{mary}T_{ij}^{sleep}$}{
 \lf{$W, T_i^{mary}$}
 \lf{$W \otimes S , T_{ij}^{sleep}$}
 }
}
\nccurve[ncurv=.6,angleA=270,angleB=160]{->}{M}{W}
\\
\end{tabular}
}\\

\end{tabular}

\end{center}
\caption{A DS with Vector Space Semantics parse of `Mary, who sleeps, snores'.}
\label{fig:linkvector}
\end{figure}

We take the DS propositional type $Ty(t)$ to correspond to a sentence space $S$, and the entity type $Ty(e)$ a word space $W$. Given vectors $T_i^{mary}, T_k^{john}$ in $W$ and  the (cube) tensor $T_{ijk}^{like}$ in $W \otimes S \otimes W$, the tensor semantic trees of the DS parsing process of $Mary \ likes \ John$  become as in Fig.~\ref{fig:marylikesjohn}.\footnote{
There has been much discussion about whether sentence and word spaces should be the same or separate. In  previous work, we have worked with both cases, i.e.\ when $W \neq S$ and when $W = S$.}

A very similar procedure is applicable to the linked structures, where conjunction can be  interpreted by the  $\mu$ map of a  Frobenius algebra over a vector space, e.g. as in  \cite{kartsaklisphd}, or as composition of the interpretations of its two conjuncts, as in \cite{MuskensSadr2016}. The $\mu$ map has also been  used to model relative clauses \cite{RelPronMoL,SadrClarkCoecke1,SadrClarkCoecke2}. It \emph{combines}  the information of the two vector spaces into one. Figure 2 shows how it combines the information of two contracted tensors $T_i^{mary}T_{ij}^{sleep}$ and $T_i^{mary}T_{ij}^{snore}$.

DS \emph{requirements} can now be treated as requirements for tensors of a particular order (e.g.\ $?W$, $?W \otimes S$ as above). If we can give these suitable vector-space representations, we can then provide an analogue to \newcite{Hough.Purver12}'s incremental type inference procedure, allowing us to compile a partial tree to specify its overall semantic representation (at its root node). One alternative would be to interpret them as picking out an element which is \emph{neutral} with regards to  composition:  the unit vector/tensor of the space they annotate. 
%
A more informative alternative would be to interpret them as enumerating all the possibilities for further development. This can be derived from all the word vector and phrase tensors of the space under question --- i.e.\  all the word and phrases whose vectors and tensors live in  $W$ and in $W \otimes S$ in this case --- by taking either the  \emph{sum} $T^{+}$ or the \emph{direct sum} $T^{\oplus}$ of these vectors/tensors. Summing will give us one  vector/tensor, accumulating the information encoded in the vectors/tensors of each word/phrase; direct summing will give us a tuple, keeping this information separate from each other. This gives us the equivalent of a sub-lattice of the record type lattices described in \cite{HoughPurver17Lattices}, with the appropriate underspecified record type as the top element, and the attendant advantages for incremental probabilistic interpretation. 

These alternatives all provide the desired compositionality, but differ in the semantic information they contribute. The use of the identity provides no semantic information; the sum gives information about the ``average'' vector/tensor expected on the basis of what is known about the language and its use in context (encoded in the vector space model); the direct sum enumerates the possibilities. In each case, more semantic information can then arrive later as more words are parsed. The best alternative will depend on task and implementation: in the next section, we give a working example using the sum operation.

\section{Incremental Plausibility: a working example}

In order to exemplify the abstract tensors and tensor  contraction operations of the model and provide a proof of concept for its applicability   to semantic incrementality,  we  characterise the incremental disambiguation of the \textit{The footballer dribbled....} example in (\ref{ex:dribble}). 
This example is worked out in the  instance of the compositional distributional  semantics introduced in \cite{SClark15} and implemented in \cite{Polajnar},  intended to model \emph{plausibility}. 
In this instance, $S$  is a two dimensional space with basis vectors true $\top$ and false $\bot$. Sentences that are highly plausible   have a vector representation close to the  $\top$ basis; highly implausible sentences  have one close to the  $\bot$ basis.   
%
%
As an illustrative example, we take $W$ to be the following $4 \times 4$ matrix based on co-occurrence counts:\footnote{For illustrative purposes, the  co-occurrence counts are taken from random excerpts of up to 100 sentences, taken from the BNC; a full implementation would of course use larger datasets.}

\begin{footnotesize}
\begin{center}
\begin{tabular}{|l|cccc|}
\hline
 & infant & nappy & pitch & goal\\
\hline
 baby & 34 & 10 & 0 & 0 \\
  milk&10 & 1 & 0 & 0 \\
  footballer&0 & 0 & 11 & 52\\
  ball&0 & 1 & 27 & 49 \\
\hline
\end{tabular}
\end{center}
\end{footnotesize}

\noindent
For an example of a vector representation,  consider the  row corresponding to \emph{baby}: this gives us a vector with the  linear expansion  $\sum_i C^{baby}_i {e}_i$,  for $e_i \in \{$\emph{infant, \ nappy, \ pitch, \ goal}$\}$ a basis vector of $W$ and $C^{baby}_i$ its corresponding scalar value. The value  $C^{baby}_{2} = 10$ represents the number of times   \emph{baby} occurred in the  same piece of text as  \emph{nappy}; the value  $C^{baby}_{4} = 0$ represents the number of times \emph{baby} occurred in the same excerpt as \emph{goal}, e.g. as the subjects of \emph{wore nappy} or \emph{crawled into a goal}. 

Intransitive verbs $v$  will have matrix representations with linear expansion $\sum_{ij} C^v_{ij} {e}_i \otimes {e}_j$  with ${e}_i$ a basis vector of $W$ and ${e}_j$ a basis vector of $S$.  A high value for  $v$ on the basis $\langle {e}_i, \top\rangle$ means that it is highly plausible that $v$   has the property $e_i$; a high value at the  $\langle {e}_i, \bot\rangle$ means  that it is highly implausible that $v$ has property $e_i$. For example, consider   the verbs \emph{vomit, score, dribble} in their intransitive roles:  $T^{score}$ has  a high value at $\langle \text{goal},\top\rangle$, since it is highly plausible that things that are scored are goals; and a high entry at  $\langle \text{nappy},\bot\rangle$, since it is highly implausible that things that wear nappies (e.g.\ babies)  score. $T^{vomit}$ has an opposite plausibility distribution for infant and nappy wearing agents.  $T^{dribble}$ is a mixture of these two, since both nappy wearing and goal scoring agents do it, but in different senses. Here, we instantiate the matrix purely from text co-occurrence, approximating plausibility from co-occurrence of verb and entity in the same text excerpt and implausibility from lack thereof, i.e.\ occurrence of verb without the entity. Other methods could of course be used, e.g.\ using dependency parse information to show verb-agent relations directly; or learning entries via regression \cite{Polajnar}. 
Note that while this makes our plausibility and implausibility degrees dependent, and the two dimensional $S$ can therefore be reduced to a one dimensional one, the theory supports spaces of any dimension, so we  present values and computations for both dimensions to illustrate this. 

\begin{footnotesize}
\begin{center}
\begin{tabular}{|l|cccccccc|}
\hline
 & 
 $\langle \text{infant} ,\top\rangle$ & 
  $\langle \text{infant} ,\bot\rangle$&
 $\langle \text{nappy} ,\top\rangle$&
  $\langle \text{nappy} ,\bot\rangle$&
  $\langle \text{pitch} ,\top\rangle$&
  $\langle \text{pitch} ,\bot\rangle$&
 $\langle \text{goal} ,\top\rangle$&
 $\langle \text{goal} ,\bot\rangle$
 \\
\hline
vomit & 10 & 2 & 9 & 3 &  3 & 9 &  0 & 12 \\
 score & 1 & 7 & 0 & 8 & 7 & 1 & 8 & 0  \\
 dribble & 22& 2 &21& 3 & 14& 10 & 16 & 8 \\
\hline
\end{tabular}
\end{center}
\end{footnotesize}


\noindent
The interpretation of an intransitive sentence, such as \textit{Babies vomit} is calculated as follows:

\[
\begin{array}{ll}
T^{\mbox{babies \ vomit}}= T_i^{babies} T_{ij}^{vomit} =&
(C^{baby}_{1}C^{vomit}_{11} + C^{baby}_{2}C^{vomit}_{21} +
C^{baby}_{3}C^{vomit}_{31} +
C^{baby}_{4}C^{vomit}_{41}) \top + \\ 
&(C^{baby}_{1}C^{vomit}_{12} + C^{baby}_{2}C^{vomit}_{22} +
C^{baby}_{3}C^{vomit}_{32} +
C^{baby}_{4}C^{vomit}_{42}) \bot\\
&= (34 \times 10 + 10 \times 9) \top + (34 \times 2 + 10 \times 3) \bot\\
&= 430 \top + 98 \bot
\end{array}
\]
Similar calculations provide us with the following sentence representations:

\begin{eqnarray*}
T^{\mbox{babies \ score}}&=& 34 \top + 318 \bot\\
T^{\mbox{babies \ dribble}}&=&  958 \top + 98 \bot\\
T^{\mbox{footballers \ vomit}} &=& 33 \top + 723 \bot \\
T^{\mbox{footballers \ score}}&=& 493 \top + 11 \bot\\
T^{\mbox{footballers \ dribble}} &=& 986 \top + 526 \bot
\end{eqnarray*}
\noindent
It follows that \textit{Babies vomit} is more plausible than \textit{Footballers vomit},  \textit{Footballers score} is more plausible than \textit{Babies score}, but \textit{Babies dribble} and \textit{Footballers dribble} have more or less the same degree of plausibility. 

A transitive verb such as \emph{control} will have a tensor representation  as follows:
\[
T^{control} = \sum_{ijk} C^{control}_{ijk} e_i \otimes e_j \otimes e_k
\]
for $e_i,e_k$ basis of $W$ and $e_j$ either $\top$ or $\bot$. Suppose that \emph{control}  has a 1 entry value at \emph{pitch} and \emph{goal} with $e_j = \top$ and a low or zero entry everywhere else.  It is easy to show that the sentence representation of \textit{Footballers control balls} is much more plausible than that of \textit{Babies control balls}.

\begin{eqnarray*}
T^{\mbox{footballers \ control \ balls}} &=& T^{{footballers}}_i T^{{control}}_{ijk} T^{{balls}}_k\\
&=& C_i^{footballer}C_k^{ball}  \langle pitch\rangle C_{ijk}^{control}(\langle pitch, \top, pitch\rangle + 
\langle pitch, \top, goal\rangle)\\
&+& C_i^{footballer} C_k^{ball} \langle goal\rangle C_{ijk}^{control}(\langle goal, \top, pitch\rangle + 
\langle goal, \top, goal\rangle)\\
&=& 6866 \top \\
T^{\mbox{babies \ control \ balls}} &=& T^{{babies}}_i T^{{control}}_{ijk} T^{{balls}}_k = 0
\end{eqnarray*}

In an unfinished utterance, such as \emph{babies \ldots}, parsing will first derive a semantic tree containing the vector for \emph{babies} and a tensor for $?W \otimes S$;  then we tensor contract the two to obtain a vector in $?S$.  The underspecified tensor in $W \otimes S$ is computed  by summing all known elements of  $W \otimes S$:
\[
T^+_{ij} = T^{vomit} + T^{score} + T^{dribble} + T^{control\  baby} + T^{control\  milk} + T^{control\  footballer} + T^{control\  ball}
\]
The tensor contraction of this with the vector of \emph{babies} provides us with the meaning of the utterance: 
\[
 T_i^{babies} T^+_{ij}
\]
%
Similar calculations to the previous cases show that plausibility increases when moving from the incomplete utterence  $T^{babies}_i T_{ij}^{+}$  to the complete one $T^{babies}_i T^{vomit}_{ij}$. Conceptually speaking, the incomplete phrase will be a dense, high-entropy vector with nearly equal values on $\top$ and $\bot$, whereas the complete phrase (or the more complete phrase), will result in a  sparser vector with more differential values on $\top$ and $\bot$. 
Continuation with a less plausible verb, e.g. \textit{score} would result in a reduction in plausibility; and different transitive verb phrases would of course have corresponding different effects. We therefore cautiously view this as an initial step towards a model which can provide the ``error  signal'' feedback assumed in models of expectation during language interpretation \cite{Clark15surfing}.

\section{Nondeterminism of Meaning and Incremental Disambiguation}

\subsection{Incremental Disambiguation}
Distributional semantics comes with a straightforward  algorithm for acquisition of word meaning, but when a word is ambiguous its vector representation becomes  a mixture of  the representations  of its  different senses. Post processing of these vectors is needed to obtain different representations for each sense  \cite{Schutze,KartSadr2013}. 
Given vectors for individual senses, our setting can incrementally disambiguate word meanings as the sentence is processed.  For instance, we can incrementally determine that in \textit{Footballers dribble}, \textit{dribble} means `control the ball'; while in 
\textit{Babies dribble}  it means something closer to `drip'. This is done by computing that \textit{Babies dribble}$_{drip}$ is more plausible than \textit{Babies dribble}$_{control}$, and also that \textit{Footballers dribble}$_{control}$  is more plausible than \textit{Footballers dribble}$_{drip}$. 
Note that this disambiguation can be made before the sentence is complete: in \textit{Her fingers tapped on her i-pad}, or \textit{The police tapped his phone}, the combination of subject and verb alone can (given suitable vectors and tensors) give information about the relative plausibility of the readings of \textit{tapped} as `knocked' or `intercepted'. This can then be strengthened when the object is parsed (or, indeed, weakened or even reversed, depending on the object).



The above examples are taken from the disambiguation dataset of \cite{Kartsaklisetal13}. Parts of this dataset has   been  tested on the plausibility model of \cite{SClark15} by \cite{Polajnar}, where it has been shown that plausibility implementations of verb tensors do a better job in disambiguating them.
Repeating this task in our model to experimentally validate  the incremental disambiguation hypothesis  constitutes  work in progress. 


\subsection{Incremental Expectation}

Using our model on examples such as the above, we can also incrementally compute plausibility of possible continuations. Consider the ``dribble'' example in (\ref{ex:dribble}) again: after parsing \textit{Footballers dribble},  we can calculate not merely that the verb's interpretation can be narrowed down in the presence of the subject, but also that the continuation \textit{ball} would be very plausible, and  the continuation \textit{milk} very implausible. 
A similar computation  provides us with the plausible continuations for \textit{Police intercept} vs \textit{Fingers knock}. If we are using the (direct) sum method to assign overall plausibility to the unfinished sentence, the plausibility values of the possible continuations have already been calculated; here we need only inspect the particular values of interest. 
Using this method, we can therefore explain how people assign shifting expectations as parsing proceeds, and make interim probabilistic evaluations on the basis thereof -- giving us a basis for a model embodying the `predictive processing' stance of \cite{Clark13}. We leave experiments into evaluating this hypothesis for future work.

\section{Discussion}

Our distributional DS model gives us a basis for incremental interpretation via compositional, grammar-driven vector space semantics. The particular instantiation outlined above assigns sentence representations in only a two-dimensional plausibility space, but the framework generalises to any vector space. Our intention is to extend this to more informative spaces, and integrate with the incremental probabilistic approaches to interpretation such as \newcite{HoughPurver17Lattices}'s approach to reference resolution.

One important step will be to adapt the model for incremental \emph{generation}. In the original formulation by \newcite{PurverKempson04Generation}, DS generation is defined as a process of DS parsing, along with a check against a \emph{goal tree}. At each generation step, every word in the vocabulary is tested to check if it is parseable from the current parse state; those which can be parsed are tested, with the resulting DS tree being checked to see if it subsumes the goal tree. If it does subsume it, then the parsed word can be generated as output; when the current tree and goal tree match, generation is complete and the process halts. \newcite{Hough.Purver12} updated this to use a goal \emph{concept} as a TTR record type, with the subsumption check now testing whether a DS-TTR tree's top-level record type is a proper supertype of (i.e.\ subsumes) the current goal record type. Given the equivalence of our proposed model to \newcite{Hough.Purver12}'s parsing process described above, the only additional apparatus required for generation for DS with Vector Space Semantics is the use of a goal \emph{tensor}, and a characterisation of subsumption between two tensors. For the latter, we intend to look into a distributional characterisation of inclusion \cite{KartsaklisSadrzadeh2016distributional}, in the spirit of a real-valued measure of relevance proposed in probabilistic type theory by \cite{HoughPurver17Lattices}. Other approaches to this are exploring type theory and vector space semantics hybrids such as \cite{asher2017types}.


\bibliographystyle{acl}
\bibliography{refs}

\end{document}